\documentclass[letterpaper]{article} 
\usepackage[preprint]{aaai2027}
\usepackage[hyphens]{url}  
\usepackage{graphicx} 
\usepackage{natbib}  
\usepackage{caption} 
\usepackage{algorithm}
\usepackage{algorithmic}
\usepackage{booktabs}
\usepackage{amsmath}
\usepackage{amssymb}
\usepackage{multirow}
\usepackage{subcaption}
\usepackage{xcolor}

\DeclareCaptionStyle{ruled}{labelfont=normalfont,labelsep=colon,strut=off}

\newcommand{\mn}[1]{\ensuremath{-}#1}

\title{A Better Start for Language Models: Domain-Conditional Position Offsets}
\author{
    Ye Qiao
}
\affiliations{
    University of California, Irvine\\
    Yeq6@uci.edu
}

\begin{document}

\maketitle

\begin{abstract}
Autoregressive language models are least accurate at the beginning of a sequence, where little context forces reliance on a generic pretraining prior. We show that this cold-start penalty is domain dependent and reduce it with a domain-conditional position offset: a single learned vector added to the embedding activation at the first sequence positions while all model weights remain frozen. The offset trains in minutes on roughly one hundred documents, switches between domains without added sequence state, and has no measurable latency overhead. Across eight Mamba, GPT-NeoX, and Llama models spanning 410M to 8B parameters, it reduces held-out in-domain perplexity by up to 27\%; the effect persists at 70B, and one position captures most of the benefit. A matched, converged direct logit-bias correction reaches at most only 7.9\% and leaves later-token loss unchanged, showing that the offset propagates through model state rather than merely recalibrating the output prior. A tuned LoRA reaches lower perplexity but uses two to three orders of magnitude more parameters and an active low-rank weight path, while soft prompts add sequence positions. With wrong-domain controls, offsets improve retrieval reranking and domain classification when decisions depend on early in-domain tokens, For the few-shot reasoning whose signal occurs later, the results maintains unchanged. Position-aware prefill application also help generation tasks, whereas naive application at every cached decoding step causes repetition. The offset is therefore not the strongest adapter, but a lightweight, hot switchable tool for short in-domain scoring and calibration.
\end{abstract}

\section{Introduction}

Every autoregressive language model faces the same difficulty at the start of a sequence: it must produce a next-token distribution with essentially no context, and it does so by falling back on an implicit prior formed during pretraining. Because that prior is an average over the entire pretraining corpus, it is rarely well matched to any particular domain, and the consequence is a measurable cold-start penalty in which early-position predictions are considerably worse than predictions made later in the same sequence, once enough context has accumulated. The penalty is not merely a curiosity of language modeling. It is paid at the beginning of every sequence, so it compounds in exactly the settings where models process many short inputs rather than one long one, such as retrieval pipelines, classification services, and multi-agent systems in which each agent invocation begins cold \citep{wu2023autogen, hong2024metagpt}.

The natural remedies carry costs for rapid, per-domain specialization. Full fine-tuning modifies the model weights, while LoRA \citep{hu2022lora} installs low-rank weight updates and requires adapter-aware serving, even though LoRA adapters can themselves be switched. Prompt and prefix tuning \citep{lester2021power, li2021prefix} prepend learned tokens that consume context positions and, in transformers, add key-value state; a plain system prompt, as we confirm, tends to move a base model out of its pretraining distribution and hurts rather than helps. We ask whether a smaller intervention can capture most of the available benefit without adding a sequence position or a weight-space adapter.

We study domain-conditional position offsets, a single learned vector, on the order of a few thousand parameters, that is added to the token embedding at the first few positions of a sequence. The vector is trained by minimizing the ordinary language modeling loss on a small sample of domain text while every model weight stays frozen, and at inference it is applied through a forward hook so that switching domains changes only one small tensor, with no added sequence state and no measurable latency in our benchmark. That a small learned prompt can reduce a language model's in-domain perplexity is already known \citep{dingliwal2021prompt}; our aim is not to re-establish that a few parameters suffice, but to characterize precisely what this intervention does and does not provide. We explain the effect as an early-position phenomenon, realize it as a positional correction that adds no token and therefore applies even to models with no start token to tune, and map where its benefit transfers. The cold-start effect and its correction reproduce cleanly across architectures and scales, and a single position carries most of the benefit, which supports the appeal of the method. At the same time, a carefully tuned LoRA reaches lower perplexity than the offset, so the contribution is one of efficiency rather than dominance, and the value of the offset lies in its cost profile: near-zero parameters, no weight modification, and no added sequence position. We then take the analysis past perplexity, since a lower loss is only useful if it changes a decision. Using a wrong-domain control that must not help if the effect is genuinely a domain prior, we find that the offset improves retrieval reranking and domain classification, both of which hinge on early in-domain tokens, while maintaining few-shot task accuracy, whose answer tokens sit far past the region the offset touches. Finally, generation task is also improved with the offset when it is applied correctly: a hook that fires at every cached decoding step degrades text into repetition, whereas restricting the offset to true early positions leaves generation quality.

Taken together, our contributions are a careful characterization of the domain-dependent cold-start penalty and its correction by a single vector across three architecture families and a wide range of scales; We compare our approach against the alternatives, including LoRA adaptation, fine-tuned start-token embedding and one-token soft prompt, that isolates what is new here, a positional correction that requires no dedicated start token, adds no additional token, and has no measurable latency; a controlled efficiency comparison against a tuned LoRA and a learned soft prefix that positions the offset as the best trade-off between cost and benefit rather than the lowest perplexity; a set of downstream results with specificity controls, including a standard classification benchmark ;and a mechanistic analysis showing, among other things, that the correction reduces a transformer's reliance on its attention sink.

\section{Related Work}

Parameter-efficient adaptation methods reduce the cost of specializing a model but modify what the offset leaves untouched. LoRA \citep{hu2022lora} injects trainable low-rank updates into weight matrices and adapters \citep{houlsby2019parameter} insert bottleneck modules between layers; both adapt effectively with far fewer parameters than full fine-tuning and can be switched, but require adapter-aware serving and active weight-space modules. Prompt tuning \citep{lester2021power} and prefix tuning \citep{li2021prefix} learn continuous vectors that are prepended to the input, which consumes context positions and, in transformers, adds key-value state during decoding. This mechanism has been applied directly to language-model domain adaptation, where a handful of prepended domain-token embeddings lower in-domain perplexity by an amount comparable to full fine-tuning \citep{dingliwal2021prompt}, an effect our own soft-prompt baseline reproduces. Related contextual-calibration work corrects label-prior biases from content-free inputs \citep{zhao2021calibrate}; our setting instead learns a domain prior for ordinary language-model scoring and localizes its effect by sequence position. The efficiency of a small learned prompt or calibrated prior is therefore not by itself new; our contribution is to identify the early-position cold-start penalty that such a correction addresses and to show that the fix need not add a token at all. Our offset differs from a prepended prompt on both axes at once: it modifies the embedding activations at the first few positions rather than lengthening the input sequence, so it adds no context position or transformer key-value state, and because it targets positions rather than a dedicated start token it applies to recurrent models such as Mamba that have no start token to tune. We treat a well-tuned LoRA and a learned soft prefix as the reference points against which the offset should be judged, and we compare against properly tuned baselines rather than undertuned ones.

A second line of work concerns the special role of the earliest tokens. Attention-sink phenomena show that transformers concentrate a large fraction of attention mass on the first token, and that preserving it is essential for stable long-context and streaming behavior \citep{xiao2024efficient}; analogously, vision transformers benefit from dedicated register tokens that absorb global information \citep{darcet2024vision}. Our method can be read as making the earliest position carry a learned, domain-specific signal rather than a generic one, which connects the cold-start penalty to this broader observation that position zero is unusually influential. State space models \citep{gu2023mamba, dao2024transformers} make the point sharper, since their recurrent computation threads the initial state through every later step, and we indeed find that they benefit more than transformers at matched scale.

Finally, the downstream setting we emphasize is scoring rather than open-ended generation. Retrieval-augmented pipelines \citep{lewis2020retrieval} and speculative decoding \citep{leviathan2023fast} both rely on a model assigning accurate likelihoods to short candidate continuations, which is precisely where an in-domain calibration at the start of a sequence can help. Continued pretraining \citep{gururangan2020dont} remains the gold standard for deep domain adaptation but is far too expensive for the rapid, hot-swappable specialization we target, and our offset is complementary to it.

\section{Method}

\subsection{The Cold-Start Bottleneck}

\begin{figure*}[t]
    \centering
    \includegraphics[width=\textwidth]{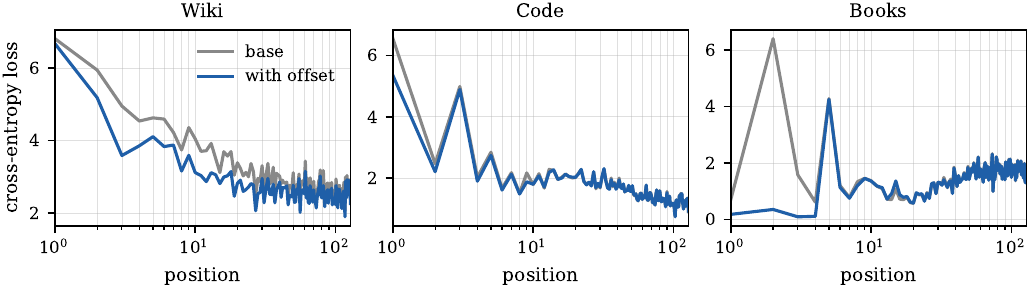}
    \caption{Per-position cross-entropy on Mamba-2.8B, with and without the domain offset, for three domains. Loss is high at the first positions and decays with context; the offset lowers loss across the early region and the benefit cascades to later positions through the model's recurrent computation.}
    \label{fig:cold-start}
\end{figure*}

Consider an autoregressive model with embedding dimension $d$. Given tokens $x_1, \ldots, x_T$, let $e_t=\texttt{Embed}(x_t)$ and let $p_t$ denote any architecture-specific positional contribution, with $p_t=0$ for architectures such as Mamba that encode order recurrently rather than through an explicit positional term. The representation fed to the first block is $h_t^{(0)}=e_t+p_t$. At $t=1$ the model must predict $x_2$ from only the first observed token, or from a dedicated beginning token when the tokenizer inserts one, so the quality of that prediction is governed by how well $h_1^{(0)}$ serves as a prior for the target domain. We hypothesize that pretraining installs a generic prior, averaged over all domains, that is suboptimal for any specific one. Measuring per-position cross-entropy confirms a pronounced and domain-dependent cold-start effect (Figure~\ref{fig:cold-start}): loss at the first content positions is several times higher than at position 64, and the size of the gap varies markedly by domain, with Wikipedia showing the strongest penalty in keeping with its high lexical specificity relative to the pretraining mixture.

\subsection{Domain-Conditional Position Offset}

We learn a domain-specific offset $\delta_\mathcal{D} \in \mathbb{R}^{K \times d}$ that is added to the embedding activations at the first $K$ positions,
\begin{equation}
    \tilde{h}_t^{(0)} = h_t^{(0)} + \delta_\mathcal{D}[t], \quad t = 1, \ldots, K,
    \label{eq:offset}
\end{equation}
while positions $t > K$ pass through unchanged. In practice a single position ($K=1$, a vector of dimension $d$) captures most of the benefit and $K$ between $3$ and $5$ adds little. For a 2.8B model with $d=2560$ the offset is between 2{,}560 and 12{,}800 parameters, a ratio of roughly $10^{-6}$ to the base model. It is applied through a forward hook on the embedding layer, with no architectural change and no custom kernels, so that in the simplest terms the hook performs $\tilde{h}[:, {:}K, :] \mathrel{+}= \delta_\mathcal{D}$.

\subsection{Training and Application}

Given $N$ documents from domain $\mathcal{D}$, we minimize the standard language modeling objective with respect to $\delta_\mathcal{D}$ alone,
\begin{equation}
    \mathcal{L}(\delta_\mathcal{D}) = -\!\!\sum_{x \in \mathcal{B}} \sum_{t} \log p_\theta\!\left(x_{t+1} \mid \tilde{h}_1, \ldots, \tilde{h}_t\right) + \lambda \lVert \delta_\mathcal{D} \rVert_2^2,
    \label{eq:loss}
\end{equation}
with a mild L2 penalty $\lambda = 0.01$. All weights $\theta$ stay frozen and gradients reach only the $K \times d$ offset parameters. We use Adam with learning rate $10^{-3}$, sequences of up to 512 tokens, and three epochs, and training completes in well under three minutes on a single GPU. Two properties make this robust. The extremely low dimensionality of the parameter space acts as a strong implicit regularizer, and because the offset shapes every later position through the model's own computation, the gradient signal it receives aggregates over the full sequence even though the vector is applied only at the first $K$ positions.

One subtlety matters for how the offset is deployed. When it is used to score existing text, or during the prompt-processing (prefill) stage of generation, the offset correctly lands on true early positions. During cached autoregressive decoding, however, each step is a length-one forward pass, so a naive hook would add $\delta_\mathcal{D}[0]$ to every newly generated token as if it were position zero. We therefore apply the offset by absolute position, adding $\delta_\mathcal{D}[p]$ only where the absolute position $p < K$, which restricts it to the prefill region and leaves ordinary decoding untouched.

\begin{algorithm}[t]
\caption{Domain-Conditional Offset}
\label{alg:offset}
\begin{algorithmic}[1]
\STATE \textbf{Train} (once per domain, under 3 min): initialize $\delta_\mathcal{D} \leftarrow \mathbf{0}$; for three epochs, for each document apply $\tilde{h}_{1:K} \leftarrow h_{1:K} + \delta_\mathcal{D}$, compute $\mathcal{L}$ (Eq.~\ref{eq:loss}) and update $\delta_\mathcal{D}$ with Adam.
\STATE \textbf{Apply} (no added overhead): register a hook that adds $\delta_{\mathcal{D}}[p]$ at absolute positions $p < K$; run the forward pass normally.
\end{algorithmic}
\end{algorithm}

\section{Experiments}

We evaluate eight publicly available pretrained models from Mamba \citep{gu2023mamba}, Pythia \citep{biderman2023pythia}, and Llama \citep{touvron2023llama2,grattafiori2024llama3}, using them without modification; the complete model inventory is in the supplementary material. Offsets are trained on Wikipedia (WikiText-103) \citep{merity2017pointer}, source code (CodeSearchNet Python) \citep{husain2019codesearchnet}, books (PG19) \citep{rae2020compressive}, and web text (C4) \citep{raffel2020exploring}, and evaluated on held-out splits of the same domains; AG News \citep{zhang2015character} provides the standard classification benchmark. Unless noted we use $K=5$, Adam at learning rate $10^{-3}$, three epochs, training sequences up to $512$ tokens, and held-out sequences up to $1024$ tokens. Table~\ref{tab:scaling} collects the original single-run sweeps, which average document-level losses; the controlled efficiency and seed studies instead aggregate token-level loss over their held-out sets. Absolute perplexities therefore should be compared only within a table, while the repeated relative reductions provide the cross-study signal. Unless a result reports an interval, it is one training run; confidence intervals use five runs with seeds 0 through 4. The matched prior-control experiment of Table~\ref{tab:prior} uses a deliberately shorter, token-weighted $512$-token evaluation to expose whether a correction propagates beyond the first positions, and the mechanism analysis restricts attention to the first $64$ positions. The confirmatory 70B run uses a reduced protocol of 150 training documents, two epochs, 120 held-out documents, and a maximum length of 320 tokens, and is therefore reported separately rather than inserted into the scaling table.

Experiments run on Linux with NVIDIA H200 GPUs, All runs use one GPU except the four-GPU 70B run.

\subsection{The Cold-Start Penalty and Its Correction}

The offset lowers in-domain perplexity for every model we tested, and the benefit grows with scale (Table~\ref{tab:scaling}). At 2.8B, the recurrent Mamba benefits about ten points more than the similarly sized Pythia, which is consistent with the view that a recurrent model leans more heavily on its initial state while a transformer can partly compensate by attending to earlier positions. The instruction-tuned Llama variant still improves by roughly a quarter, indicating that instruction tuning does not by itself remove the cold-start mismatch. The values in Table~\ref{tab:scaling} are from a single training run per model. To confirm that the effect does not hinge on the random seed, we retrain the Wikipedia offset five times for four models under a fixed protocol of 300 documents. The reduction is stable across seeds: at $K=5$ it is $21.5 \pm 0.7$\% for Mamba-2.8B, $20.0 \pm 0.4$ for Llama-3.2-3B, $20.7 \pm 2.0$ for Llama-2-7B, and $23.5 \pm 4.5$ for Llama-3.1-8B, and a single position already realizes most of it, for example $19.8 \pm 0.2$\% on Mamba-2.8B and $22.0 \pm 1.2$ on Llama-3.1-8B, where the ranges are 95\% confidence intervals over the seeds. The effect also holds at the largest scale we could run: sharding Llama-3.1-70B across four GPUs, the $K=5$ offset lowers held-out Wikipedia perplexity by $23.3$\%, from $5.70$ to $4.37$, so the correction does not wash out as the base model grows.

\begin{table}[t]
    \centering
    \caption{Held-out Wikipedia perplexity reduction by model ($K=5$) in the original single-run sweeps, using sequences up to 1024 tokens and document-level loss aggregation. Absolute PPL is shown for context; relative reduction is the intended cross-model comparison.}
    \label{tab:scaling}
    \resizebox{\columnwidth}{!}{%
    \begin{tabular}{llccc}
        \toprule
        Model & Family & Base PPL & Offset PPL & $\Delta$\% \\
        \midrule
        Pythia-410M & GPT-NeoX & 34.71 & 31.87 & \mn{8.2} \\
        Mamba-790M & SSM & 23.62 & 20.13 & \mn{14.8} \\
        Pythia-2.8B & GPT-NeoX & 21.43 & 17.96 & \mn{16.2} \\
        Mamba-2.8B & SSM & 18.86 & 13.90 & \mn{26.3} \\
        Llama-3.2-3B & Llama & 16.10 & 12.44 & \mn{22.7} \\
        Llama-2-7B & Llama & 10.17 & 7.59 & \mn{25.4} \\
        Llama-3.1-8B & Llama & 13.58 & 9.82 & \textbf{\mn{27.7}} \\
        Llama-3.1-8B-Inst & Llama & 14.88 & 11.31 & \mn{24.0} \\
        \bottomrule
    \end{tabular}%
    }
\end{table}

\paragraph{Direct prior correction.}
Because our account invokes a mismatched domain prior, a direct output-space correction is an especially important control. We therefore train three logit-bias baselines on the same 500 documents as fresh $K=1$ and $K=5$ offsets: one vocabulary bias at the first position, one shared vocabulary bias at the first five positions, and separate biases at each of those positions. We sweep learning rates from $0.01$ to $1.0$ and train for up to 100 epochs with validation-based checkpointing and early stopping; all selected checkpoints are interior to both search bounds. All methods are evaluated token-wise on the same 200 documents at length 512. Table~\ref{tab:prior} shows that the strongest direct bias reduces full perplexity by $0.6$\% on Mamba and $7.9$\% on Llama, despite using 50K to 641K parameters, while the embedding offset reduces it by $23$ to $24$\% with 2.6K to 20K parameters. More importantly, every logit bias leaves loss after position five exactly unchanged, whereas the offset lowers tail perplexity by $21.0$\% on Mamba and $17.5$\% on Llama. The gain is therefore not merely a better unigram prior at the corrected logits: changing the early representation propagates through the model state.

\begin{table}[t]
        \centering
        \caption{Matched output-prior control on Wikipedia (500 train/200 test documents, 512 tokens). We report full-sequence perplexity change and tail perplexity change after position five. Under teacher-forced scoring, direct logit biases leave the tail unchanged; the embedding offset improves both.}
        \label{tab:prior}
        \resizebox{\columnwidth}{!}{%
        \begin{tabular}{lrrcc}
                \toprule
                Model & Method & Params & Full $\Delta$\% & Tail $\Delta$\% \\
                \midrule
                \multirow{5}{*}{Mamba-2.8B}
                      & Logit bias, $K{=}1$ & 50.3K & \mn{0.02} & $0.0$ \\
                      & Shared logit bias, $K{=}5$ & 50.3K & \mn{0.6} & $0.0$ \\
                    & Position logit bias, $K{=}5$ & 251K & \mn{0.4} & $0.0$ \\
                      & Offset $K{=}1$ & 2.6K & \mn{22.2} & \mn{21.0} \\
                      & Offset $K{=}5$ & 12.8K & \textbf{\mn{23.5}} & \textbf{\mn{22.5}} \\
                \midrule
                \multirow{5}{*}{Llama-3.1-8B}
                      & Logit bias, $K{=}1$ & 128K & \mn{7.9} & $0.0$ \\
                      & Shared logit bias, $K{=}5$ & 128K & \mn{3.0} & $0.0$ \\
                      & Position logit bias, $K{=}5$ & 641K & \mn{6.2} & $0.0$ \\
                    & Offset $K{=}1$ & 4.1K & \mn{23.3} & \mn{14.3} \\
                    & Offset $K{=}5$ & 20.5K & \textbf{\mn{24.3}} & \textbf{\mn{17.5}} \\
                \bottomrule
        \end{tabular}%
        }
\end{table}

\subsection{A Controlled Efficiency Comparison}
\label{sec:ladder}

The obvious question is why not simply use LoRA. Answering it fairly requires tuning LoRA rather than running it once at a fixed learning rate, since an undertuned LoRA overfits a few hundred documents and its perplexity can explode with rank, which would flatter the offset for the wrong reason. We therefore sweep the LoRA learning rate over $\{10^{-4}, 3{\times}10^{-4}, 10^{-3}\}$ with early stopping on a validation split and report the best configuration per rank, and we add a learned soft prefix as a second strong baseline. Table~\ref{tab:ladder} gives the resulting frontier, with a visualization in the supplementary material. A well-tuned LoRA is stable across rank and reaches the lowest perplexity, roughly a third on Mamba and a quarter to a bit more on Llama. The offset does not match that number, but it recovers about 75\% of it with two to three orders of magnitude fewer parameters, without touching any weight and without adding a sequence position or transformer key-value state. The soft prefix is not only perform worse but pays a per-step cost because it lengthens the sequence. In short, the offset is not the strongest adapter; it is the cheapest useful one, and it is the only option on this frontier that is both weight-free and free of added sequence positions.

\begin{table}[t]
    \centering
    \caption{Parameter-efficiency frontier on Wikipedia. A tuned LoRA reaches the lowest perplexity; the offset recovers most of the benefit at a tiny fraction of the parameters, with no weight change or added sequence position. A soft prefix lengthens the sequence and, in transformers, adds kv-cache.}
    \label{tab:ladder}
    \resizebox{\columnwidth}{!}{%
    \begin{tabular}{lrrcc}
        \toprule
        Method & Params & $\Delta$\% & Weights & Extra pos. \\
        \midrule
        \multicolumn{5}{l}{\textit{Mamba-2.8B, base PPL 15.82}} \\
        \quad Offset $K{=}1$ & 2{,}560 & \mn{21.6} & frozen & 0 \\
        \quad Offset $K{=}5$ & 12{,}800 & \mn{22.2} & frozen & 0 \\
        \quad Soft prefix $M{=}5$ & 12{,}800 & \mn{14.7} & frozen & 5 \\
        \quad LoRA $r{=}1$ (tuned) & 819\,K & \mn{31.7} & modified & 0 \\
        \quad LoRA $r{=}4$ (tuned) & 3.3\,M & \mn{31.7} & modified & 0 \\
        \quad LoRA $r{=}16$ (tuned) & 13.1\,M & \mn{31.5} & modified & 0 \\
        \midrule
        \multicolumn{5}{l}{\textit{Llama-3.1-8B, base PPL 11.17}} \\
        \quad Offset $K{=}1$ & 4{,}096 & \mn{21.0} & frozen & 0 \\
        \quad Offset $K{=}5$ & 20{,}480 & \mn{15.7} & frozen & 0 \\
        \quad Soft prefix $M{=}5$ & 20{,}480 & \mn{15.1} & frozen & 5 \\
        \quad LoRA $r{=}1$ (tuned) & 426\,K & \mn{25.1} & modified & 0 \\
        \quad LoRA $r{=}4$ (tuned) & 1.7\,M & \mn{26.0} & modified & 0 \\
        \quad LoRA $r{=}16$ (tuned) & 6.8\,M & \mn{27.9} & modified & 0 \\
        \bottomrule
    \end{tabular}%
    }
\end{table}

\subsection{Comparison to Baseline Alternatives}
\label{sec:minimal}

A single vector added at the first position invites two objections: that it is nothing more than a one-token soft prompt, or nothing more than a fine-tuned start-of-sequence embedding. Table~\ref{tab:minimal} addresses both. Fine-tuning the embedding of the start token is indeed similar to our $K{=}1$ offset when such a token exists. The equivalence then breaks in two informative ways. Mamba has no dedicated start token, so fine-tuning a start embedding is simply inapplicable and leaves perplexity unchanged, whereas the positional offset still reduces it by more than a fifth; the offset is the positional generalization of a start-token embedding and is defined for every architecture. The offset also extends to positions $1$ through $K{-}1$, which hold a different token on every sequence and therefore cannot be written as any edit to the token-embedding table. A one-token learned soft prompt not only have higher perplexity than the offset, but it also occupies a sequence position, adds transformer key-value state during decoding, and adds a recurrent step on Mamba, whereas a training-free vector built from the mean domain embedding does not help at all, so the benefit comes from the learned direction rather than from any domain-shifted vector.

Additionally, we also measure the cost difference. We time a single scoring forward pass, the operation that underlies re-ranking and classification, at batch sixteen (Table~\ref{tab:cost}). Since the offset only adds a precomputed vector to a handful of embeddings through a forward hook, it creates no new sequence position, and its latency is indistinguishable from the unmodified model at every length, within run-to-run noise below 1\%, with identical peak memory. The apples-to-apples one-token prompt has modest but measurable overhead on Llama-3.1-8B, adding $4.1$\% at length 16 and $2.0$\% at length 64, while five prepended tokens add $15.0$ and $8.1$\%. At length 512 the one-token overhead falls to $0.1$\%, so we do not claim a material latency advantage for long inputs. On Mamba, prepended tokens add recurrent steps.

\begin{table}[t]
    \centering
    \caption{Runtime cost of a single scoring forward pass at batch sixteen, as the change in latency relative to the unmodified model. The offset adds no sequence position and is indistinguishable from the base model; soft-prompt overhead is modest for one token and largest on short inputs.}
    \label{tab:cost}
    \resizebox{\columnwidth}{!}{%
    \begin{tabular}{lrccc}
        \toprule
        Length & Base (ms) & Offset $K{=}5$ & Soft $M{=}1$ & Soft $M{=}5$ \\
        \midrule
        \multicolumn{5}{l}{\textit{Llama-3.1-8B}} \\
        \quad 16 & 13.5 & $+0.2\%$ & $+4.1\%$ & $+15.0\%$ \\
        \quad 64 & 32.4 & $\mn{0.7}\%$ & $+2.0\%$ & $+8.1\%$ \\
        \quad 256 & 122.5 & $\mn{0.2}\%$ & $+2.8\%$ & $+3.5\%$ \\
        \multicolumn{5}{l}{\textit{Mamba-2.8B}} \\
        \quad 16 & 32.7 & $+0.1\%$ & $+1.8\%$ & $+4.6\%$ \\
        \quad 32 & 37.0 & $\mn{0.1}\%$ & $+2.3\%$ & $+5.8\%$ \\
        \bottomrule
    \end{tabular}
    }
\end{table}

\begin{table}[t]
    \centering
    \caption{The offset against minimal alternatives (Wikipedia, perplexity change). A start-token embedding equals the $K{=}1$ offset where a start token exists (Llama) but is inapplicable without one (Mamba); a one-token soft prompt is stronger yet adds a sequence position; a training-free mean vector does not help.}
    \label{tab:minimal}
    \resizebox{\columnwidth}{!}{%
    \begin{tabular}{lccccc}
        \toprule
        Method & Params & Mamba & Llama & Added pos. & Start tok. \\
        \midrule
        Mean embedding (train-free) & $0$ & $+0.1$ & $+3.5$ & 0 & no \\
        Start-token embedding & $d$ & $+0.0$ & \mn{22.1} & 0 & needs \\
        1-token soft prompt & $d$ & \mn{21.2} & \mn{20.9} & 1 & no \\
        \textbf{Offset $K$ = 1 (ours)} & $d$ & \textbf{\mn{22.3}} & \textbf{\mn{22.1}} & \textbf{0} & \textbf{no} \\
        \textbf{Offset $K$ = 5 (ours)} & $5d$ & \textbf{\mn{23.9}} & \mn{16.6} & \textbf{0} & \textbf{no} \\
        \bottomrule
    \end{tabular}%
    }
\end{table}

\subsection{Domain Specificity}

The offset encodes a direction, not merely a magnitude. Table~\ref{tab:cross-domain} gives the cross-domain matrix for Mamba-2.8B: the matched offset improves perplexity on its own domain, while a mismatched offset hurts, and the harm is largest for the most lexically specific domains. This specificity is what makes the offset usable as a discriminator in the next section, and it also motivates selecting the right offset at deployment time. A lightweight character $n$-gram router reaches $94.7$ to $98.7$\% accuracy across the four domains; routed perplexity stays close to the oracle choice, including reductions of $24.8$ versus $26.3$\% on Wikipedia and $3.2$ versus $3.3$\% on web text.

\begin{table}[t]
    \centering
    \caption{Cross-domain perplexity change (\%) for Mamba-2.8B. The diagonal is the matched, in-domain offset; off-diagonal entries apply a mismatched offset and degrade performance.}
    \label{tab:cross-domain}
    \resizebox{\columnwidth}{!}{%
    \begin{tabular}{lcccc}
        \toprule
        Eval $\downarrow$ / Train $\rightarrow$ & Wiki & Code & Books & Web \\
        \midrule
        Wikipedia & \textbf{\mn{26.3}} & +4.4 & +0.2 & +1.2 \\
        Code & +14.2 & \textbf{\mn{1.6}} & +1.9 & +0.1 \\
        Books & +5.9 & +1.0 & \textbf{\mn{1.9}} & +2.0 \\
        Web & +15.9 & +3.6 & +0.9 & \textbf{\mn{3.3}} \\
        \bottomrule
    \end{tabular}%
    }
\end{table}

\subsection{Beyond Perplexity: Downstream Tasks}
\label{sec:downstream}

A lower loss is only useful if it changes a decision, so we test the offset on two tasks that turn its in-domain calibration into a native metric and that depend on early, in-domain tokens. In retrieval reranking we form a query from the first tokens of a document and ask the model to rank the true continuation against distractors drawn from other documents, scoring each candidate by its likelihood given the query; in domain classification we label a short text by which domain offset assigns it the lowest loss. Crucially, both include a wrong-domain control: if a mismatched offset helped as much as the matched one, the effect would not be attributable to a domain prior. Table~\ref{tab:downstream} shows this pattern most clearly on Wikipedia, the domain with the strongest cold-start penalty, where the in-domain offset raises reranking accuracy by 7\% to 13\% on both models while the wrong-domain offset degrade the performance. On code and books, whose cold-start penalty is far smaller, the effect is correspondingly small but still obvious. Classification, by contrast, is near perfect for Llama and above 92\% for Mamba using only three small vectors as the entire classifier.

\begin{table}[t]
    \centering
    \caption{Discriminative downstream tasks with a wrong-domain specificity control. Reranking reports accuracy@1 as the mean and 95\% confidence interval over five seeds on the in-domain column; classification reports accuracy over three domains on 16-token to 128-token inputs. On Wikipedia the in-domain offset helps and the wrong-domain offset does not, isolating the effect to the domain prior.}
    \label{tab:downstream}
    \resizebox{\columnwidth}{!}{%
    \begin{tabular}{llccc}
        \toprule
        \multicolumn{5}{l}{\textit{Reranking accuracy@1}} \\
        Model & Domain & Base & In-domain & Wrong \\
        \midrule
        \multirow{3}{*}{Mamba-2.8B} & Wiki & 0.702 & \textbf{0.792}\,{\footnotesize$\pm$.01} & 0.706 \\
         & Code & 0.672 & \textbf{0.692}\,{\footnotesize$\pm$.01} & 0.646 \\
         & Books & 0.472 & \textbf{0.494}\,{\footnotesize$\pm$.03} & 0.466 \\
        \multirow{3}{*}{Llama-3.1-8B} & Wiki & 0.650 & \textbf{0.726}\,{\footnotesize$\pm$.05} & 0.622 \\
         & Code & 0.700 & \textbf{0.752}\,{\footnotesize$\pm$.03} & 0.640 \\
         & Books & 0.376 & \textbf{0.397}\,{\footnotesize$\pm$.06} & 0.364 \\
        \midrule
        \multicolumn{5}{l}{\textit{Domain classification accuracy (3-way)}} \\
        Model & 16 tok & 32 tok & 64 tok & 128 tok \\
        \midrule
        Mamba-2.8B & 0.924 & 0.938 & 0.953 & 0.935 \\
        Llama-3.1-8B & 1.000 & 1.000 & 1.000 & 1.000 \\
        \bottomrule
    \end{tabular}%
    }
\end{table}

The same mechanism explains a negative result. Few-shot accuracy on standard benchmarks does not move: with two hundred items per task the standard error is around three points and every difference we observe falls inside it. This is expected rather than disappointing, because a few-shot answer token sits far past the demonstrations, in the region the offset never touches, and the benchmark content is out of the offset's domain. The offset helps precisely when the decision depends on the first in-domain tokens, and not otherwise.

To test classification on a standard benchmark rather than three deliberately distinct domains, we apply the same likelihood classifier to AG News, a four-way news-topic task, training one offset per topic on a few hundred headlines. With roughly fifty thousand parameters in total serving as the entire classifier, and no trained classification head, it reaches $74.7$\% accuracy for Mamba-2.8B and $70.5$\% for Llama-3.1-8B against a chance rate of $25$\%, and accuracy is highest on the shortest inputs, consistent with the offset acting on the earliest tokens. For reference, a supervised linear probe on mean-pooled final-layer hidden states, which trains a dedicated classification head and reads the model's full internal representation, reaches $84.2$ and $84.5$\% on the two models. The offset classifier does not match a trained probe, which is unsurprising since it carries no head and decides purely by per-class likelihood, but it recovers well over 80\% of the probe's accuracy above chance with four small vectors and no supervised readout, which is the sense in which the perplexity effect is strong enough to be read out as a decision.

\begin{figure*}[t]
    \centering
    \begin{subfigure}[t]{0.68\textwidth}
        \centering
        \includegraphics[width=\textwidth]{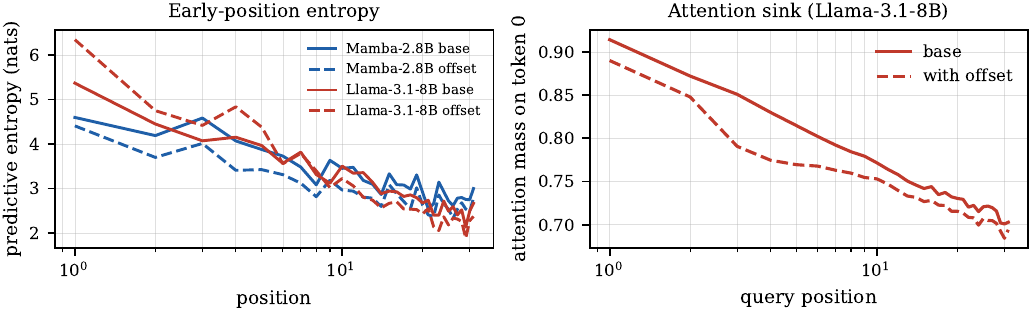}
        \phantomsubcaption
        \label{fig:mech}
    \end{subfigure}
    \hfill
    \begin{subfigure}[t]{0.3\textwidth}
        \centering
        \includegraphics[width=\textwidth]{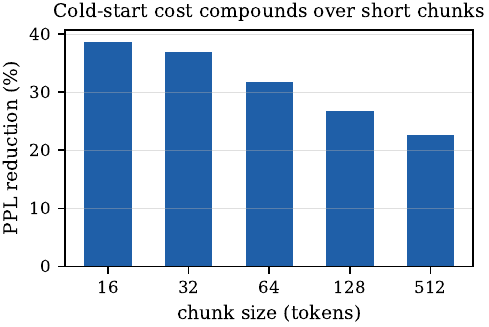}
        \phantomsubcaption
        \label{fig:comp}
    \end{subfigure}
    \caption{Mechanism and compounding. (a) Early predictive entropy and Llama attention-sink mass. (b) Mamba perplexity reduction grows as independent chunks shorten.}
    \label{fig:analysis}
\end{figure*}

\subsection{Generation}

Deploying the offset during generation requires care, because its effect interacts with key-value caching. In the prefill stage the offset lands on the true early positions, but during cached decoding each step is a length-one forward pass, and a hook that fired at every step would add $\delta_\mathcal{D}[0]$ to every generated token as though it were at position zero. This constant embedding shift accumulates across the output and drives the model into a repetition attractor. The position-aware application of Algorithm~\ref{alg:offset}, which adds $\delta_\mathcal{D}[p]$ only at absolute positions $p<K$, confines the offset to the prefill region and avoids this behavior entirely. Table~\ref{tab:gen} contrasts the two: the per-step application collapses diversity and inflates repetition, whereas the position-aware application make both better than baseline auto regressive generation. To measure quality without allowing the offset to flatter its own output, we score the generated text under an independent referee model that was never adapted to any domain; under this referee the position-aware generation is statistically better from the baseline on Mamba and and Llama. Generation is thus fully compatible and performs slightly better with our proposed offset, even if it is not where the method contributes most.

\begin{table}[t]
    \centering
    \caption{Generation under greedy decoding, aggregated over prompts, domains, and both models (ranges shown). The naive per-step application degenerates; the position-aware (prefill-only) application surpass the baseline by a small margin.}
    \label{tab:gen}
    \begin{tabular}{lcc}
        \toprule
        Application & distinct-2 $\uparrow$ & repeat-4gram $\downarrow$ \\
        \midrule
        Baseline (no offset) & 0.75 to 0.96 & 0.00 to 0.13 \\
        Naive (every step) & 0.04 to 0.66 & 0.05 to 0.96 \\
       \textbf{ Position-aware (ours)} & \textbf{0.81 to 0.98} & \textbf{0.00 to 0.09} \\
        \bottomrule
    \end{tabular}
\end{table}

\subsection{Ablations}

Two ablations characterize the operating point. A single position captures nearly all of the benefit: on Mamba-2.8B, $K=1$ reaches within a point of the best $K$, and larger $K$ eventually regresses slightly, which locates the bottleneck almost entirely at the sequence start. The method is also data efficient: one hundred documents and about half a minute of training already deliver roughly half of the maximum improvement, and the gain grows smoothly toward its plateau near five hundred documents, so a new domain can be equipped with an offset in minutes.

\section{Analysis}

\subsection{Mechanism}

Two probes clarify how a vector at the first positions improves later predictions. Figure~\ref{fig:mech} (left) reports the model's predictive entropy at each early position with and without the offset, and the two architectures respond differently. On the recurrent Mamba the offset lowers early-position entropy by about 11\%, sharpening otherwise diffuse initial predictions, whereas on Llama it slightly raises entropy, tempering an overconfident initial distribution; in both cases the effect is to move the early distribution toward the target token. For the transformer we can say more. Figure~\ref{fig:mech} (right) measures the fraction of attention placed on the first token, the attention sink that transformers rely on for stability~\citep{xiao2024efficient}. The base model routes the large majority of early-position attention onto token zero, and the offset reduces this reliance by giving the early positions a more informative representation of their own, so that later positions lean less on the sink. The offset can therefore be read as installing a learned, domain-specific signal at the very position a transformer already treats as special, which also connects the cold-start penalty to the register tokens found useful in vision transformers~\citep{darcet2024vision}.

To test whether the sink response tracks the correction, we scale the trained Llama offset and measure, over the first sixty-four positions, both perplexity and attention placed on token zero. Strengthening the offset reduces sink reliance monotonically, from $0.857$ at the base model to $0.798$ at twice the trained magnitude. Perplexity instead falls to a minimum and then rises: it drops by 42\% at about three-quarters of the trained magnitude and remains 38\% lower at the trained value, before climbing as the offset is pushed further. The two quantities move together throughout the beneficial regime, establishing that the offset controls sink reliance, but they diverge beyond the optimum, showing that sink reduction alone is not sufficient to explain lower perplexity. This is consistent with the sink's stabilizing role: the useful intervention reduces over-reliance without eliminating it.

\subsection{Compounding over Short Sequences}

Because the penalty is paid at the start of every sequence, its cost accumulates when a corpus is processed as many short segments rather than one long one. Figure~\ref{fig:comp} makes this concrete for the recurrent architecture: processing held-out text in independent chunks, the perplexity reduction grows from about 23\% at five-hundred-token chunks to 39\% at sixteen-token chunks, since each restart pays the cold-start cost that the offset removes. This is the regime of many short in-domain sequences in which a near-free correction is most valuable, and it is strongest for the recurrent model whose state must be rebuilt from scratch at every restart.

\section{Discussion and Limitations}

The method's value is its cost profile rather than peak accuracy: tuned LoRA reaches lower perplexity, while the offset remains weight-free, adds no sequence state, and shows no measurable latency. Its locality is also its boundary: it helps early-token scoring tasks but not few-shot reasoning whose signal occurs later. Speculative-decoding gains are mixed rather than robust. Our systematic sweep stops at 8B, with one reduced-protocol confirmation at 70B, and covers four language-modeling domains plus AG News; multilingual and specialized-domain evaluation remains future work. Gains are necessarily small when the base model's initial prior already matches the target domain.

\section{Conclusion}

Cold start is not a model-specific defect but a structural condition of autoregressive language modeling: every model must make its first predictions before informative context exists, exposing a generic pretraining prior that may be poorly matched to the current domain. We show that this mismatch can be corrected at the point where it begins. Across eight Mamba, GPT-NeoX, and Llama models from 410M to 8B parameters, plus a 70B confirmation, domain-conditional position offsets with only 2.6K to 40.9K trainable parameters reduce held-out in-domain perplexity by up to 27\%; five-seed reductions on 2.8B to 8B models remain 20.0\% to 23.5\%, compared with at most 7.9\% from a converged direct logit-bias correction. The offset modifies no model weights, adds no token or transformer key-value state, and incurs no measurable latency. Its effect propagates beyond the corrected positions, improves Wikipedia reranking accuracy by 7 to 9 points, and leaves late-token reasoning unchanged; position-aware application also avoids the repetition collapse caused by naive cached decoding and text generation performance also improved by small margin. Together, these results establish domain-dependent cold start as a general consequence of autoregressive prediction and reframe the model's initial state as an actionable adaptation interface, opening a distinct and broadly applicable direction for adapting language models without changing their weights.

\clearpage
\appendix
\setcounter{secnumdepth}{1}

\section{Evaluated Models}

\begin{table}[t]
    \centering
    \caption{Public pretrained models evaluated in the main paper.}
    \label{tab:arxiv-models}
    \resizebox{\columnwidth}{!}{%
    \begin{tabular}{llcc}
        \toprule
        Model & Architecture & Params & $d$ \\
        \midrule
        Pythia-410M & GPT-NeoX & 410M & 1024 \\
        Mamba-790M & SSM (Mamba) & 790M & 1536 \\
        Pythia-2.8B & GPT-NeoX & 2.8B & 2560 \\
        Mamba-2.8B & SSM (Mamba) & 2.8B & 2560 \\
        Llama-3.2-3B & Llama & 3B & 3072 \\
        Llama-2-7B & Llama & 7B & 4096 \\
        Llama-3.1-8B & Llama & 8B & 4096 \\
        Llama-3.1-8B-Instruct & Llama & 8B & 4096 \\
        \bottomrule
    \end{tabular}%
    }
\end{table}

\section{Parameter-Efficiency Frontier}

\begin{figure*}[t]
    \centering
    \includegraphics[width=0.80\textwidth]{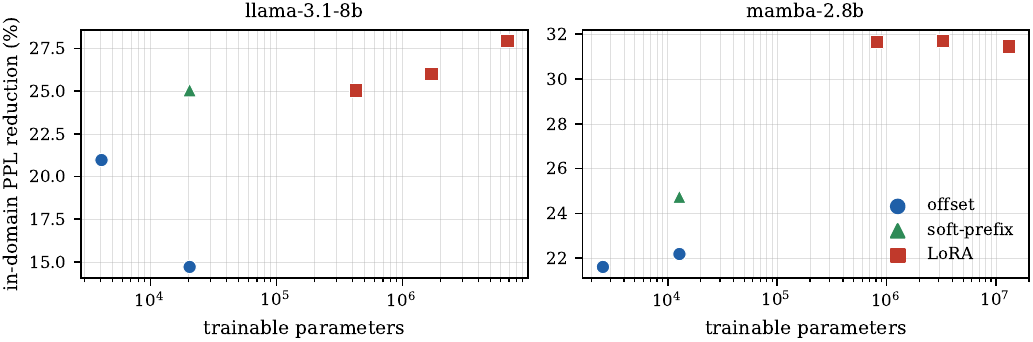}
    \caption{Parameter-efficiency frontier on Wikipedia. In-domain perplexity reduction is plotted against trainable parameters on a logarithmic scale for the offset, a learned soft prefix, and tuned LoRA adapters.}
    \label{fig:arxiv-frontier}
\end{figure*}

Figure~\ref{fig:arxiv-frontier} visualizes the controlled comparison reported in the main paper: tuned LoRA reaches the largest reduction, the offset approaches it at a much smaller parameter budget, and the soft prefix occupies an intermediate point while lengthening every input.

\section{Operating-Point Ablations}

Table~\ref{tab:arxiv-ablation} gives the position-count and data-efficiency ablations: one position captures nearly all of the gain, and performance approaches its plateau near five hundred documents.

\begin{table}[t]
    \centering
    \caption{Left: position-count ablation on Mamba-2.8B and Wikipedia. Right: training-data efficiency.}
    \label{tab:arxiv-ablation}
    \begin{tabular}{rcc@{\hskip 1.5em}rcc}
        \toprule
        $K$ & Params & $\Delta$\% & Docs & Time & $\Delta$\% \\
        \midrule
        1 & 2{,}560 & \mn{24.6} & 10 & 9s & \mn{0.2} \\
        3 & 7{,}680 & \mn{25.8} & 50 & 19s & \mn{3.3} \\
        5 & 12{,}800 & \mn{25.6} & 100 & 32s & \mn{11.5} \\
        10 & 25{,}600 & \mn{25.4} & 200 & 64s & \mn{18.3} \\
        20 & 51{,}200 & \mn{24.3} & 500 & 143s & \mn{24.2} \\
        \bottomrule
    \end{tabular}
\end{table}


\clearpage
\begin{thebibliography}{23}
\providecommand{\natexlab}[1]{#1}

\bibitem[{Biderman et~al.(2023)Biderman, Schoelkopf, Anthony, Bradley, O'Brien,
  Hallahan, Khan, Purohit, Prashanth, Raff, Skowron, Sutawika, and Van
  Der~Wal}]{biderman2023pythia}
Biderman, S.; Schoelkopf, H.; Anthony, Q.~G.; Bradley, H.; O'Brien, K.;
  Hallahan, E.; Khan, M.~A.; Purohit, S.; Prashanth, U.~S.; Raff, E.; Skowron,
  A.; Sutawika, L.; and Van Der~Wal, O. 2023.
\newblock Pythia: A Suite for Analyzing Large Language Models Across Training
  and Scaling.
\newblock In \emph{Proceedings of the 40th International Conference on Machine
  Learning}, volume 202, 2397--2430.

\bibitem[{Dao and Gu(2024)}]{dao2024transformers}
Dao, T.; and Gu, A. 2024.
\newblock Transformers are {SSMs}: Generalized Models and Efficient Algorithms
  Through Structured State Space Duality.
\newblock In \emph{Proceedings of the 41st International Conference on Machine
  Learning}, volume 235, 10041--10071.

\bibitem[{Darcet et~al.(2024)Darcet, Oquab, Mairal, and
  Bojanowski}]{darcet2024vision}
Darcet, T.; Oquab, M.; Mairal, J.; and Bojanowski, P. 2024.
\newblock Vision Transformers Need Registers.
\newblock In \emph{International Conference on Learning Representations}.

\bibitem[{Dingliwal et~al.(2021)Dingliwal, Shenoy, Bodapati, Gandhe, Gadde, and
  Kirchhoff}]{dingliwal2021prompt}
Dingliwal, S.; Shenoy, A.; Bodapati, S.; Gandhe, A.; Gadde, R.~T.; and
  Kirchhoff, K. 2021.
\newblock Prompt-tuning in {ASR} systems for efficient domain-adaptation.
\newblock \emph{arXiv preprint arXiv:2110.06502}.

\bibitem[{Grattafiori et~al.(2024)Grattafiori, Dubey, Jauhri, Pandey, Kadian,
  Al-Dahle, Letman, Mathur, Schelten, Vaughan et~al.}]{grattafiori2024llama3}
Grattafiori, A.; Dubey, A.; Jauhri, A.; Pandey, A.; Kadian, A.; Al-Dahle, A.;
  Letman, A.; Mathur, A.; Schelten, A.; Vaughan, A.; et~al. 2024.
\newblock The {Llama 3} Herd of Models.
\newblock \emph{arXiv preprint arXiv:2407.21783}.

\bibitem[{Gu and Dao(2023)}]{gu2023mamba}
Gu, A.; and Dao, T. 2023.
\newblock Mamba: Linear-Time Sequence Modeling with Selective State Spaces.
\newblock \emph{arXiv preprint arXiv:2312.00752}.

\bibitem[{Gururangan et~al.(2020)Gururangan, Marasovi{\'c}, Swayamdipta, Lo,
  Beltagy, Downey, and Smith}]{gururangan2020dont}
Gururangan, S.; Marasovi{\'c}, A.; Swayamdipta, S.; Lo, K.; Beltagy, I.;
  Downey, D.; and Smith, N.~A. 2020.
\newblock Don't Stop Pretraining: Adapt Language Models to Domains and Tasks.
\newblock In \emph{Proceedings of the 58th Annual Meeting of the Association
  for Computational Linguistics}, 8342--8360.

\bibitem[{Hong et~al.(2024)Hong, Zhuge, Chen, Zheng, Cheng, Wang, Zhang, Wang,
  Yau, Lin, Zhou, Ran, Xiao, Wu, and Schmidhuber}]{hong2024metagpt}
Hong, S.; Zhuge, M.; Chen, J.; Zheng, X.; Cheng, Y.; Wang, J.; Zhang, C.; Wang,
  Z.; Yau, S. K.~S.; Lin, Z.; Zhou, L.; Ran, C.; Xiao, L.; Wu, C.; and
  Schmidhuber, J. 2024.
\newblock {MetaGPT}: Meta Programming for A Multi-Agent Collaborative
  Framework.
\newblock In \emph{International Conference on Learning Representations}.

\bibitem[{Houlsby et~al.(2019)Houlsby, Giurgiu, Jastrzebski, Morrone,
  De~Laroussilhe, Gesmundo, Attariyan, and Gelly}]{houlsby2019parameter}
Houlsby, N.; Giurgiu, A.; Jastrzebski, S.; Morrone, B.; De~Laroussilhe, Q.;
  Gesmundo, A.; Attariyan, M.; and Gelly, S. 2019.
\newblock Parameter-Efficient Transfer Learning for {NLP}.
\newblock In \emph{Proceedings of the 36th International Conference on Machine
  Learning}, volume~97, 2790--2799.

\bibitem[{Hu et~al.(2022)Hu, Shen, Wallis, Allen-Zhu, Li, Wang, Wang, and
  Chen}]{hu2022lora}
Hu, E.~J.; Shen, Y.; Wallis, P.; Allen-Zhu, Z.; Li, Y.; Wang, S.; Wang, L.; and
  Chen, W. 2022.
\newblock {LoRA}: Low-Rank Adaptation of Large Language Models.
\newblock In \emph{International Conference on Learning Representations}.

\bibitem[{Husain et~al.(2019)Husain, Wu, Gazit, Allamanis, and
  Brockschmidt}]{husain2019codesearchnet}
Husain, H.; Wu, H.-H.; Gazit, T.; Allamanis, M.; and Brockschmidt, M. 2019.
\newblock {CodeSearchNet} Challenge: Evaluating the State of Semantic Code
  Search.
\newblock \emph{arXiv preprint arXiv:1909.09436}.

\bibitem[{Lester, Al-Rfou, and Constant(2021)}]{lester2021power}
Lester, B.; Al-Rfou, R.; and Constant, N. 2021.
\newblock The Power of Scale for Parameter-Efficient Prompt Tuning.
\newblock In \emph{Proceedings of the 2021 Conference on Empirical Methods in
  Natural Language Processing}, 3045--3059.

\bibitem[{Leviathan, Kalman, and Matias(2023)}]{leviathan2023fast}
Leviathan, Y.; Kalman, M.; and Matias, Y. 2023.
\newblock Fast Inference from Transformers via Speculative Decoding.
\newblock In \emph{Proceedings of the 40th International Conference on Machine
  Learning}, volume 202, 19274--19286.

\bibitem[{Lewis et~al.(2020)Lewis, Perez, Piktus, Petroni, Karpukhin, Goyal,
  K{\"u}ttler, Lewis, Yih, Rockt{\"a}schel, Riedel, and
  Kiela}]{lewis2020retrieval}
Lewis, P.; Perez, E.; Piktus, A.; Petroni, F.; Karpukhin, V.; Goyal, N.;
  K{\"u}ttler, H.; Lewis, M.; Yih, W.-t.; Rockt{\"a}schel, T.; Riedel, S.; and
  Kiela, D. 2020.
\newblock Retrieval-Augmented Generation for Knowledge-Intensive {NLP} Tasks.
\newblock In \emph{Advances in Neural Information Processing Systems},
  volume~33, 9459--9474.

\bibitem[{Li and Liang(2021)}]{li2021prefix}
Li, X.~L.; and Liang, P. 2021.
\newblock Prefix-Tuning: Optimizing Continuous Prompts for Generation.
\newblock In \emph{Proceedings of the 59th Annual Meeting of the Association
  for Computational Linguistics and the 11th International Joint Conference on
  Natural Language Processing (Volume 1: Long Papers)}, 4582--4597.

\bibitem[{Merity et~al.(2017)Merity, Xiong, Bradbury, and
  Socher}]{merity2017pointer}
Merity, S.; Xiong, C.; Bradbury, J.; and Socher, R. 2017.
\newblock Pointer Sentinel Mixture Models.
\newblock In \emph{International Conference on Learning Representations}.

\bibitem[{Rae et~al.(2020)Rae, Potapenko, Jayakumar, Hillier, and
  Lillicrap}]{rae2020compressive}
Rae, J.~W.; Potapenko, A.; Jayakumar, S.~M.; Hillier, C.; and Lillicrap, T.~P.
  2020.
\newblock Compressive Transformers for Long-Range Sequence Modelling.
\newblock In \emph{International Conference on Learning Representations}.

\bibitem[{Raffel et~al.(2020)Raffel, Shazeer, Roberts, Lee, Narang, Matena,
  Zhou, Li, and Liu}]{raffel2020exploring}
Raffel, C.; Shazeer, N.; Roberts, A.; Lee, K.; Narang, S.; Matena, M.; Zhou,
  Y.; Li, W.; and Liu, P.~J. 2020.
\newblock Exploring the Limits of Transfer Learning with a Unified Text-to-Text
  Transformer.
\newblock \emph{Journal of Machine Learning Research}, 21(140): 1--67.

\bibitem[{Touvron et~al.(2023)Touvron, Martin, Stone, Albert, Almahairi,
  Babaei, Bashlykov, Batra, Bhargava, Bhosale et~al.}]{touvron2023llama2}
Touvron, H.; Martin, L.; Stone, K.; Albert, P.; Almahairi, A.; Babaei, Y.;
  Bashlykov, N.; Batra, S.; Bhargava, P.; Bhosale, S.; et~al. 2023.
\newblock {Llama 2}: Open Foundation and Fine-Tuned Chat Models.
\newblock \emph{arXiv preprint arXiv:2307.09288}.

\bibitem[{Wu et~al.(2023)Wu, Bansal, Zhang, Wu, Li, Zhu, Jiang, Zhang, Zhang,
  Liu et~al.}]{wu2023autogen}
Wu, Q.; Bansal, G.; Zhang, J.; Wu, Y.; Li, B.; Zhu, E.; Jiang, L.; Zhang, X.;
  Zhang, S.; Liu, J.; et~al. 2023.
\newblock {AutoGen}: Enabling Next-Gen {LLM} Applications via Multi-Agent
  Conversation.
\newblock \emph{arXiv preprint arXiv:2308.08155}.

\bibitem[{Xiao et~al.(2024)Xiao, Tian, Chen, Han, and
  Lewis}]{xiao2024efficient}
Xiao, G.; Tian, Y.; Chen, B.; Han, S.; and Lewis, M. 2024.
\newblock Efficient Streaming Language Models with Attention Sinks.
\newblock In \emph{International Conference on Learning Representations}.

\bibitem[{Zhang, Zhao, and LeCun(2015)}]{zhang2015character}
Zhang, X.; Zhao, J.; and LeCun, Y. 2015.
\newblock Character-level Convolutional Networks for Text Classification.
\newblock In \emph{Advances in Neural Information Processing Systems},
  volume~28.

\bibitem[{Zhao et~al.(2021)Zhao, Wallace, Feng, Klein, and
  Singh}]{zhao2021calibrate}
Zhao, Z.; Wallace, E.; Feng, S.; Klein, D.; and Singh, S. 2021.
\newblock Calibrate Before Use: Improving Few-shot Performance of Language
  Models.
\newblock In \emph{Proceedings of the 38th International Conference on Machine
  Learning}, volume 139, 12697--12706.

\end{thebibliography}
\end{document}